RESEARCH NOTE

# Surprise as a Signal for Plasticity and Metacognition

Two proof-of-concept systems built on a frozen-backbone surprise detector

Louis Mouchon

ABSTRACT

We study a single idea across two settings: that a prediction-error signal, computed by a small predictor over the latent space of a frozen encoder, can serve both as a gate on plasticity and as a substrate for metacognition. In the first system, a non-parametric episodic memory writes a new concept only when this surprise is high, and a periodic offline replay phase consolidates recent traces into a slow linear readout. On a continual stream of 1000 ImageNet classes with a frozen DINOv2 or I-JEPA backbone, the consolidation phase recovers 17.7 points of retention on the oldest classes for DINOv2 and 51.3 points for I-JEPA (single-seed runs), and an ablation shows that replaying only a recent window is worse than no replay at all. In few-shot evaluation the same memory reaches 91.6% on 5-way 1-shot mini-ImageNet, above a task-specific baseline, while a harder 500-way regime exposes the true difficulty. In the second system, the same surprise signal, computed in a shared text-image space, modulates the behaviour of a vision-language model: it answers assertively when a concept is known, hedges when it is partially familiar, and refuses to identify the object and asks for an explanation when it is novel, learning the concept from a single user utterance. The external detector separates known from novel concepts at an AUROC of 0.966 (95% CI $\pm$0.024), far above the model's own verbalised confidence (0.618), while its token-level confidence sits below chance under greedy decoding; after a sleep phase that empties the fast store, the system recalls 99.2% of fifty taught facts from the consolidated store while a base model recovers none. We report both systems as proof-of-concept, with explicit limitations, and position the second against recent episodic-memory and personalised-VLM work.

## Table of contents

# Introduction

A frozen self-supervised encoder produces a useful but static representation: it knows how to embed an image or a sentence, but it has no notion of whether a given input is familiar or new. Yet that distinction is exactly what a learning agent needs in order to decide *when* to write to memory, *how much* to adapt, and *how confident* to be when it answers. In the brain, this role is played by prediction error and by neuromodulatory gating: a surprising stimulus triggers plasticity, a familiar one does not, and the hippocampus and neocortex divide the labour between fast episodic capture and slow semantic consolidation [1], [2].

This note follows a single thread through two systems, which together we call SGM (Surprise-Gated Memory). The thread is that a small predictor trained over the latent space of a frozen backbone, in the spirit of a Joint-Embedding Predictive Architecture (JEPA) [3], yields a *surprise* signal, and that this one signal is enough to drive two qualitatively different behaviours. In the first system the surprise gates plasticity: a non-parametric episodic store writes a concept only when surprise is high, and an offline replay phase consolidates it into a slow parametric readout. In the second system the surprise drives metacognition: a vision-language model reads its own novelty score and changes register, from assertive to questioning, refusing to name an object it has never been taught and learning it from a single explanation.

Neither system fine-tunes the backbone. The backbone stays frozen throughout, which by construction rules out catastrophic forgetting at the representation level and sidesteps the loss-of-plasticity failure mode that afflicts networks trained continually with gradient descent [4]. All adaptation happens in light, separable modules on top.

We make three contributions. First, we show that a JEPA-style surprise signal, calibrated against the latent geometry of a frozen encoder, separates known from novel inputs with an AUROC of 0.93 to 1.0 and that a simple threshold turns this into an effective write gate. Second, we measure the value of an offline consolidation phase on a 1000-class continual stream and find that its benefit grows with scale and with the difficulty of the backbone features, reaching +51.3 points for I-JEPA; the same set of ablations shows that a sliding-window replay is actively harmful. Third, we transfer the surprise signal to a conversational vision-language model and obtain an anti-hallucination behaviour driven by an *external* detector rather than by the model's own internally estimated confidence: the external detector separates known from novel concepts at an AUROC of 0.966 where the model's own token-level confidence falls below chance, and after a sleep phase that empties the fast store the system still recalls 99.2% of fifty taught facts. The same mechanism supports single-shot concept learning and a memory that overrides the model's prior when the two disagree.

We report both systems as proof-of-concept. The benchmarks are small and close to saturation, and the comparisons are not strictly like-for-like since the backbone is pretrained rather than trained from scratch. We state these limitations explicitly in the Limitations section rather than in a footnote.

# Background

### Joint-embedding prediction and frozen features

A JEPA learns by predicting the representation of one part of an input from another, in latent space, rather than reconstructing pixels [3]. The byproduct we exploit is the *prediction error* itself. When a small predictor is trained to recover the full embedding of a frozen encoder from a masked view, its residual is low on inputs whose structure it has seen and high on inputs it has not. We use DINOv2 [5] and I-JEPA [3] as frozen image backbones, and SigLIP [6] as a shared text-image backbone for the second system. The text retrieval channel uses BGE-M3 [7].

### Complementary learning systems and consolidation

Complementary Learning Systems theory holds that intelligent agents need a fast-learning hippocampal store with sparse, pattern-separated traces and a slow-learning neocortical store that interleaves new and old experience during offline replay [1], [2]. The machine learning analogues are experience replay and generative replay for continual learning [8], and regularisation methods such as Elastic Weight Consolidation that penalise change to important weights [9]. Our first system instantiates the two-store structure directly, with a non-parametric hippocampus and a linear neocortex, and uses replay during a discrete sleep phase.

### Test-time memory and episodic memory for language models

A separate line of work makes the memory itself adapt at inference. Titans learns to memorise at test time by taking gradient steps on a neural memory, with an update gated by a surprise term defined as the gradient of the memory's associative loss with respect to the input [10]. Episodic-memory controllers for language models such as Larimar add a one-shot writable store for knowledge editing [11], and agent memory systems such as MemGPT manage a tiered context without an offline consolidation cycle [12]. We deliberately keep our fast store non-parametric rather than test-time-trained, because a prototype written once is immune to the internal forgetting that a continually updated parametric memory suffers when it stops rewriting an old concept; we expand on this choice in the System 1 section.

### Few-shot classification and personalised vision-language models

Metric-learning few-shot classifiers such as Prototypical Networks [13], Matching Networks [14] and DeepEMD [15] learn an embedding in which class prototypes separate. With a strong frozen backbone the prototype recipe becomes highly competitive in the standard 5-way regime, which is why we also report a 500-way regime. On the language side, MyVLM [16] and Yo'LLaVA [17] personalise a vision-language model to user-specific concepts from a handful of images. Neither uses an external novelty gate, a surprise score with a refusal option, or a memory that overrides the model; these are the axes on which our second system differs, and we return to them in the Positioning section.

## System 1: surprise-gated plasticity

### Architecture

The first system is a continual image classifier built entirely on top of a frozen encoder. Figure 1 shows the data path. An input image is embedded by the frozen backbone. A small JEPA anchor, a predictor that recovers the full embedding from a masked view under a smooth-L1 loss on L2-normalised features, produces a scalar surprise equal to its residual. A non-parametric hippocampus stores recent embeddings, keyed by their surprise: a new trace is written only when surprise exceeds a threshold, so a well-known class leaves the memory untouched while a novel one is captured. A slow linear neocortex is the actual classifier. It is updated only during a periodic sleep phase that replays the hippocampal buffer, interleaving old and new traces, after which the buffer is trimmed to a small core.

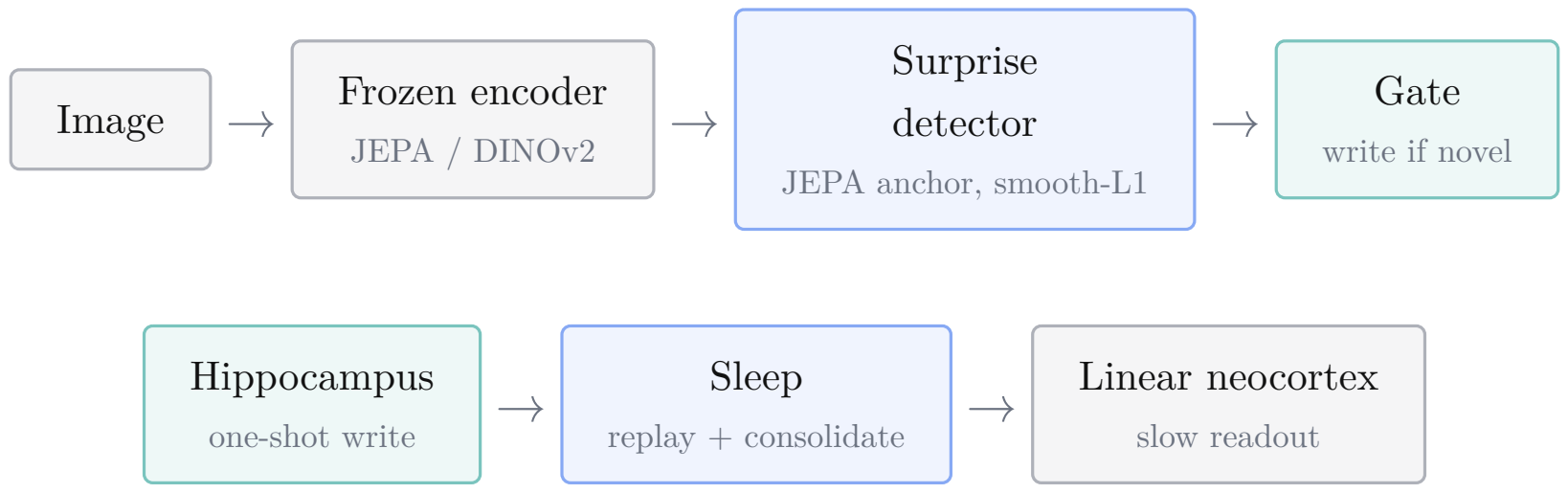

*High surprise → hippocampal write; periodic sleep → migration to neocortex (CLS).*

Figure 1: System 1. A frozen backbone feeds a JEPA surprise detector that gates writes to a fast non-parametric hippocampus; a periodic sleep phase replays and consolidates traces into a slow linear neocortex.

The backbone is never modified. The only parametric component that learns is the linear neocortex, and it learns only from replayed embeddings, never by touching the encoder. This is what makes the system immune to representation-level forgetting and to loss of plasticity.

### The surprise signal is well-calibrated

Before using surprise as a gate we checked that it actually separates known from novel inputs. We trained the anchor on a set of base classes and measured, on held-out instances, the area under the ROC curve for distinguishing base classes from unseen classes using the surprise score. The detector reaches an AUROC between 0.93 and 1.0 across base-set sizes from 10 to 400 classes. The raw separation in absolute terms is modest, but the ranking is near-perfect: a novel input almost always scores higher than a base input. A non-linear gate, with a threshold calibrated on the base distribution, converts this ranking into a clean write decision, with an operating point of F1 = 0.91 between writing on novel inputs and not writing on known ones.

### Few-shot classification

We first evaluate the memory as a few-shot classifier. In the standard 5-way regime on mini-ImageNet, prototypes built from k support embeddings of the frozen DINOv2 backbone reach 91.6% at 1-shot and 97.6% at 5-shot, above the DeepEMD-Sampling task-specific baseline at 68.8% and 84.1% [15]. With I-JEPA features the same recipe reaches 74.1% and 89.3%. The comparison is not like-for-like (DeepEMD trains its encoder from scratch while ours is pretrained on far more data), so this reflects the strength of the frozen backbone as much as the method: a strong pretrained encoder makes 5-way few-shot nearly trivial.

To expose the real difficulty we also run a 500-way regime, where each of 500 novel classes must be told apart from the other 499. Here DINOv2 prototypes reach 51.6% at 1-shot, 72.2% at 5-shot and 74.3% at 10-shot, against a chance level of 0.2%. The 500-way numbers are unchanged whether the anchor is calibrated on 30 or on 200 examples per base class, which indicates that the detector saturates early and that the few-shot behaviour depends on the support set, not on heavy base-set training. Figure 2 contrasts the two regimes.

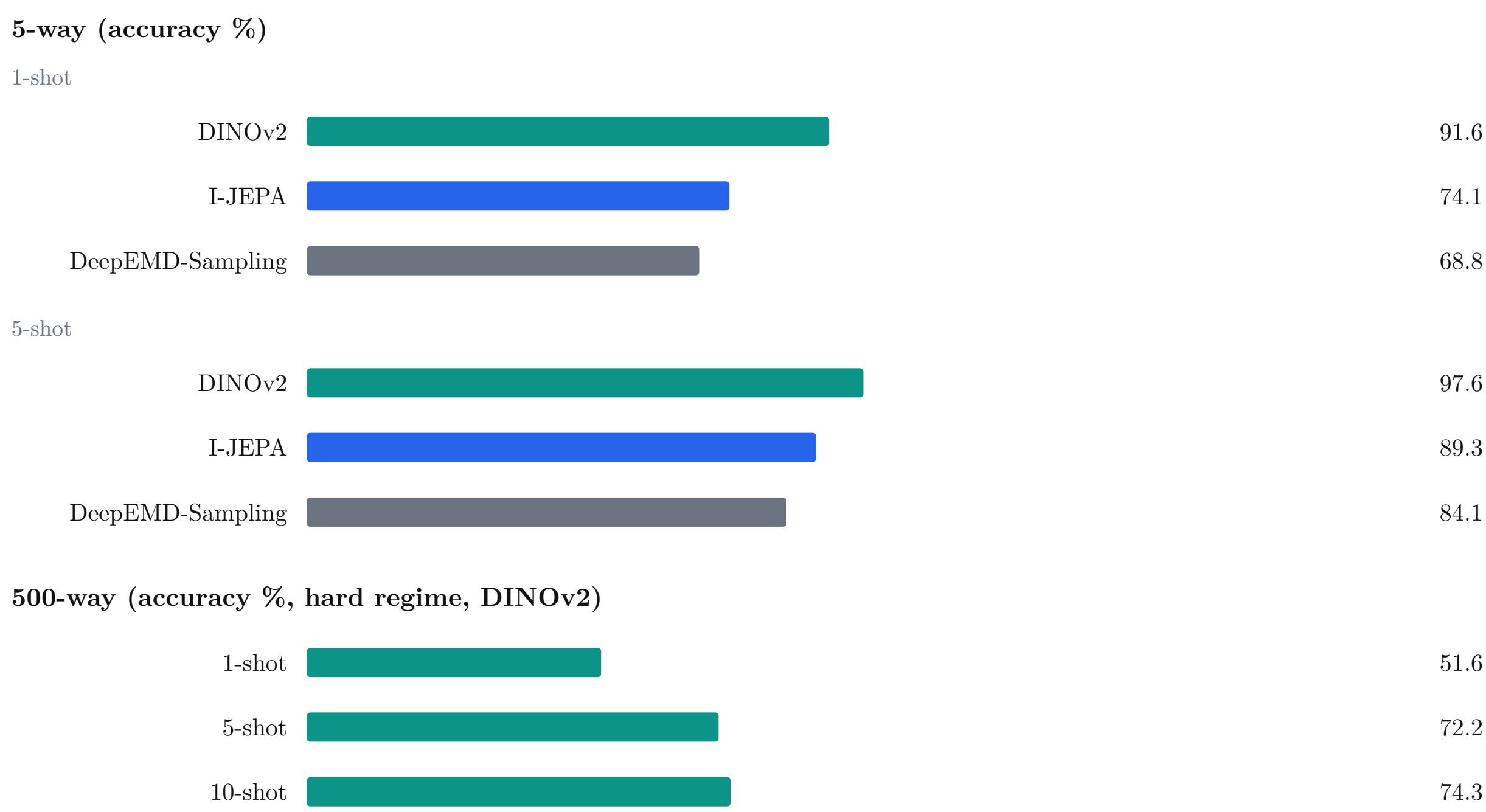

In 5-way the memory exceeds the task-specific baseline (DeepEMD). In 500-way the problem is hard but tractable and sample-efficient (identical with 30 or 200 base examples).

Figure 2: Few-shot accuracy. In the 5-way regime the frozen-backbone prototype memory exceeds a task-specific baseline; the 500-way regime is where the problem is actually hard.

### Online continual learning at 1000 classes

We then run the full system online. The classifier is pre-consolidated on a set of innate classes, then a continuous mixed stream of innate and novel classes arrives. For each image the system perceives, optionally writes to the hippocampus when surprise is high, and periodically sleeps. On a stream spanning 1000 ImageNet classes the system reaches 74.0% with DINOv2 and 59.6% with I-JEPA at the end of the stream, against a chance level of 0.1%. The backbone is never touched, so the older classes are not erased at the representation level; what is at stake is whether the slow readout retains them.

### Sleep is the load-bearing component

To isolate the contribution of the consolidation phase we compare three conditions on a stream of 1000 classes presented as 50 sequential tasks of 20 classes, and measure the retention of the five oldest tasks at the end. A naive condition trains the readout task by task with no replay; a sleep condition replays the full interleaved buffer at each step; an oracle trains on all classes i.i.d. Figure 3 reports the result. For DINOv2 the naive readout retains 65.8% of the oldest classes, sleep retains 83.5%, and the oracle ceiling is 84.4%: replay recovers 17.7 of the 18.6 available points. For I-JEPA, whose features are less linearly separable and which therefore suffers far more from interference, the naive readout collapses to 25.9% while sleep recovers it to 77.2%, against an oracle of 78.5%: a gain of 51.3 points. The benefit of consolidation is not a fixed offset; it grows with scale and with how unfriendly the backbone features are.

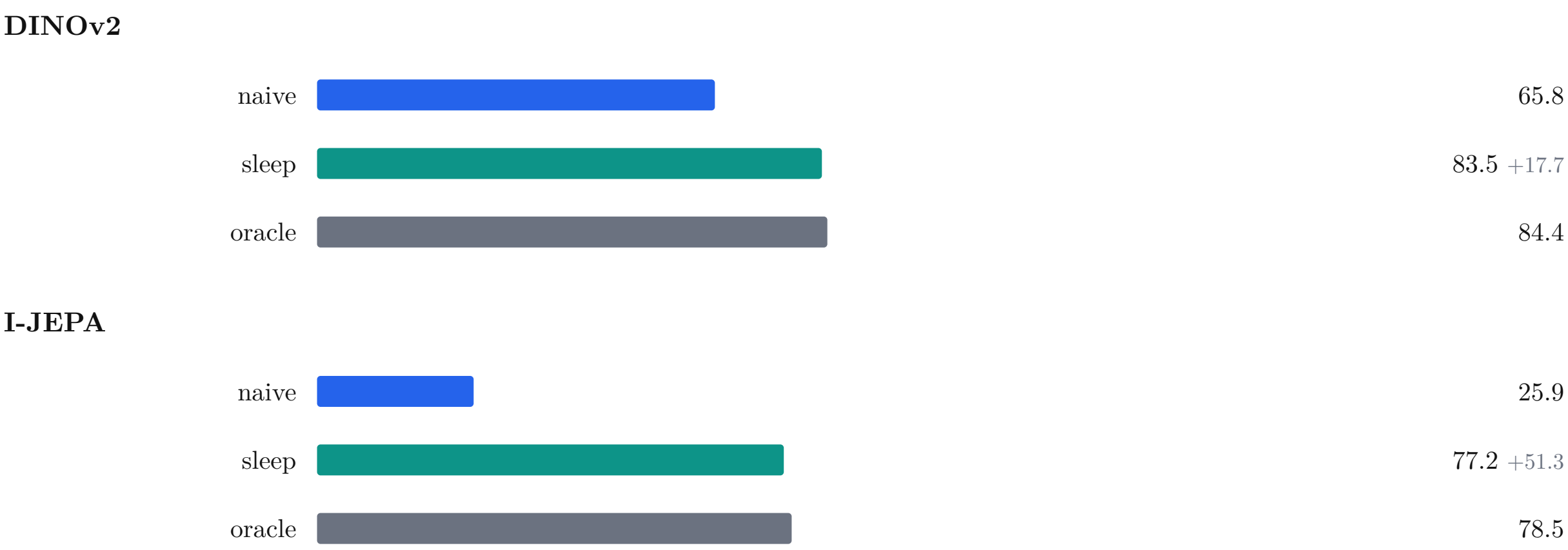

Figure 3: Retention of the oldest classes after 50 sequential tasks at 1000 classes. Sleep recovers most of the gap to the i.i.d. oracle, and the recovered margin is far larger for the harder I-JEPA backbone.

### Ablations of the memory

Figure 4 ablates the consolidation memory at 1000 classes; these are separate runs from Figure 3, so the full-replay and oracle numbers differ from the sleep and oracle numbers there by a few tenths of a point. The most informative comparison is between full replay and a sliding-window replay that only rehearses the last five tasks. The windowed variant retains only 41.2% of the oldest classes for DINOv2, and 0.0% for I-JEPA (the I-JEPA windowed result is reported in text only; Figure 4 charts the DINOv2 ablation), which is worse than the naive baseline: rehearsing only recent material actively displaces the old. Full replay reaches 83.7%. Writing only high-surprise examples, which halves the stored count, matches full replay at 84.2%, a difference well within the single-run variation we cannot rule out, so the surprise gate buys a compression of the buffer at no measurable cost rather than an improvement. A tiny memory of three examples per class already recovers most of the benefit at 78.8%. Finally, replacing the linear readout with an MLP degrades retention, because the extra capacity overfits the small replay buffer and forgets more violently between tasks; in this regime the simplicity of a linear readout is an advantage, not a limitation.

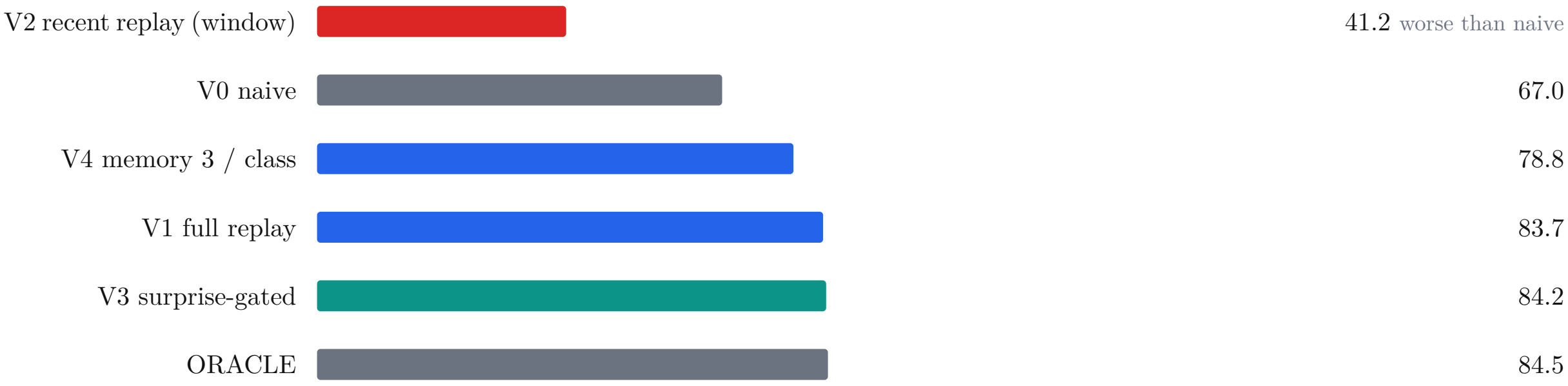

Figure 4: Memory ablations at 1000 classes (DINOv2). A sliding-window replay (V2) is worse than no replay at all. Surprise-gated writing (V3) matches full replay (V1) at half the storage.

### Why a non-parametric store

We keep the fast store non-parametric by design. A continually updated parametric memory, whether a test-time-trained neural memory or a regularisation-protected readout, forgets whatever it stops rewriting: writes triggered by novel inputs erode the weights that encode old, rarely revisited concepts. A prototype written once and never overwritten does not have this failure mode. We therefore let the hippocampus hold inert traces and let the slow neocortex, refreshed only by interleaved replay during sleep, carry the entire parametric load. This keeps the two failure modes separated: the fast store cannot forget, and the slow store forgets only as much as imperfect replay allows, which the ablations above bound directly.

# System 2: surprise-driven metacognition for a vision-language model

### From a gate to a modulator

The second system reuses the surprise signal, but instead of a binary write gate it becomes a continuous modulator of a vision-language model's behaviour. The encoder is now SigLIP [6], which places text and images in a shared space, so that the same novelty computation applies to both modalities. A textual fact retrieval channel uses BGE-M3 [7], which is far better than SigLIP at matching a question to a stored fact phrased differently. The speaker is a vision-language model, either Qwen2.5-VL 7B [18] or Gemma 4 12B [19]. Figure 5 shows the full path.

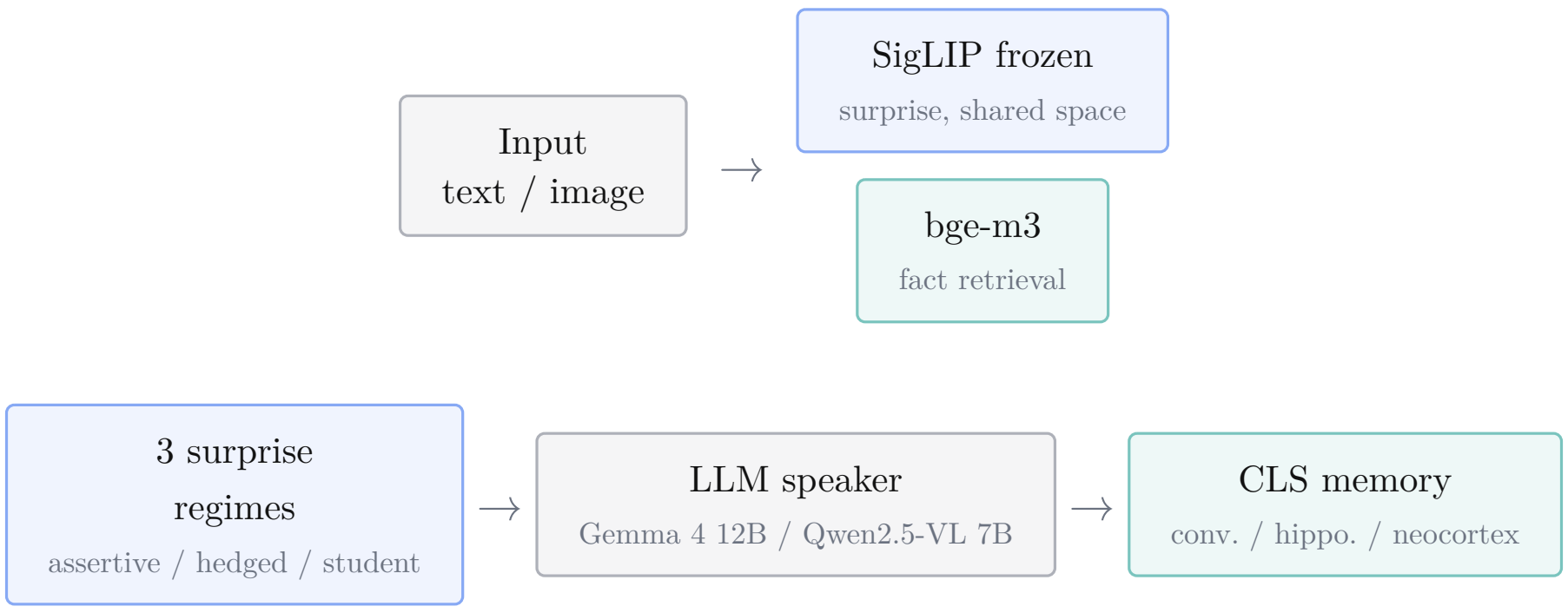

*The stored fact has authority and corrects the model hallucination. Sleep loop: hippocampus → neocortex.*

Figure 5: System 2. SigLIP computes a novelty score and BGE-M3 retrieves facts; the score selects one of three behavioural regimes injected into the prompt of a frozen vision-language model, with a three-tier memory and a sleep cycle.

The surprise here is one minus a z-scored similarity: we calibrate the distribution of cosine similarities between unrelated concepts and express the best match to memory as a deviation from that baseline. This matters because raw SigLIP similarities are compressed, with unrelated pairs around 0.72 rather than near zero; without calibration a genuinely novel image scored 0.39 and slipped into the partially-familiar regime, where the model began to hallucinate. After calibration a known concept scores about 0.14 and a novel one about 1.0, and the regime boundaries at 0.35 and 0.65 become meaningful.

### Three behavioural regimes

The novelty score selects one of three registers, written into the system prompt of the vision-language model (Figure 6). Below 0.35 the model is told the concept is known and answers assertively from the retrieved fact. Between 0.35 and 0.65 it hedges, proposing an identification but flagging its uncertainty. Above 0.65 it takes the posture of a student: it describes only what it perceives, states plainly that the concept is new to it, and asks the user to explain it. The decision comes from the *external* SigLIP detector, not from the model's own self-reported confidence, so the model cannot talk itself into naming an object that the detector has flagged as novel. This is the anti-hallucination mechanism.

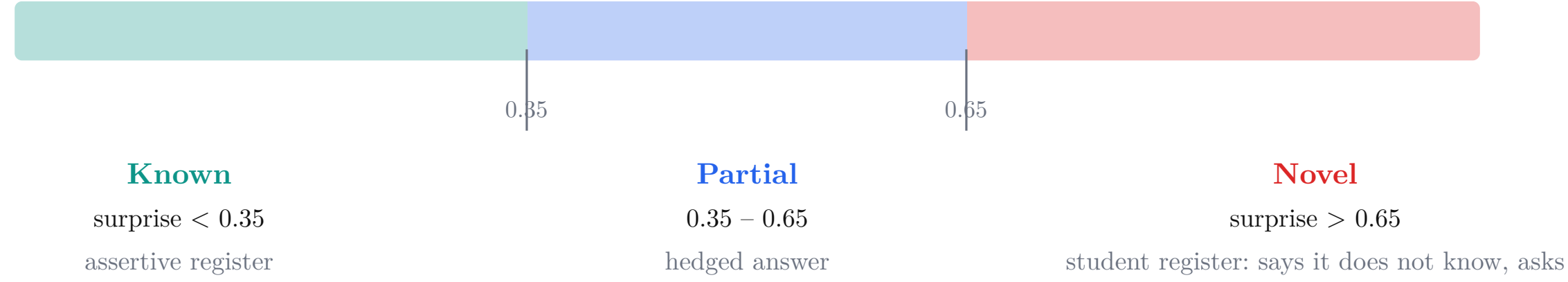

*Calibrated surprise: known ~ 0.14, novel ~ 1.0.*

Figure 6: Behavioural regimes as a function of the external novelty score. The thresholds at 0.35 and 0.65 separate assertive, hedged and student registers.

The reason to put the novelty decision in a separate detector rather than in the generator is that the generator's own confidence is unreliable in exactly the regime that matters. We measured this directly. From a heterogeneous pool of concepts (people, places, technical items, invented terms), we drew, per seed, fifteen known concepts taught to the system and fifteen never-taught novel ones, and asked three detectors to separate known from novel: our external detector (the bge surprise score), the model's verbalised confidence (asking it to rate 0–100 how sure it is it knows the answer), and the model's token-level confidence (the probability mass it puts on "yes, I know" versus "no"). Over eight such draws the external detector reaches an AUROC of 0.966 (95% CI ±0.024), and its confidence interval does not overlap the generator's verbalised confidence at 0.618; the token-level confidence stays at 0.292, *below chance*, because the model asserts it knows the invented concepts as confidently as the real ones (Figure 7). This is the confident-hallucination failure mode, measured: the generator cannot be trusted to flag its own ignorance, which is precisely why the gate is external. The gap is not an artefact of greedy decoding: under stochastic sampling the external AUROC is unchanged (the detector produces no text), while the verbalised confidence degrades further, to 0.51 (chance), confirming that the model's self-assessment is not even stable, let alone reliable.

**Known-vs-novel separation (AUROC; 15 known + 15 novel resampled per seed, 8 draws)**

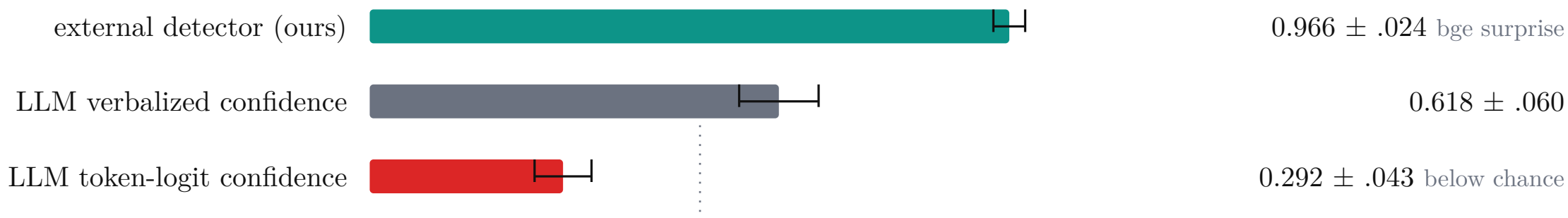

Each query is about a known (taught) or a never-taught novel concept, drawn from a heterogeneous pool (people, places, technical, fictional) and resampled per seed; error bars are 95% CIs over eight draws. The external frozen detector separates known from novel at 0.966, and its CI does not overlap the generator's verbalised confidence (0.618); the token-logit confidence stays **below chance** (0.292), as the model asserts knowledge of invented concepts. Dotted line = chance (0.5).

Figure 7: Separating known from novel concepts. The external frozen detector (AUROC 0.966, 95% CI ±0.024) outperforms the generator's own verbalised confidence (0.618) with non-overlapping confidence intervals; the generator's token-level confidence is below chance (0.292), as it asserts knowledge of invented concepts. Fifteen known and fifteen novel concepts resampled per seed from a heterogeneous pool, eight draws; error bars are 95% CIs.

### One-shot learning and authority of memory

When a concept is novel and the user explains it, the system stores the pair of the input embedding and the explanation text in the hippocampus. The explanation is captured automatically: a short classification call asks the model whether the user's message states a fact worth remembering, so the user does not need a keyword. The next time the concept appears, the novelty score is low, the fact is retrieved, and the model answers from it. The retrieved fact is given *authority*: it is injected with an instruction that it overrides the model's own prior, including negations. This was necessary because the vision-language model, shown a learned abstract image, would otherwise describe it from raw appearance and contradict what it had just been taught; with the authority instruction it reconciles the two, naming the concept from memory while acknowledging the appearance. The same mechanism lets a user correct a factual error, for instance the altitude of a landmark, and have the correction take precedence over the model's pretrained but wrong belief.

### A three-tier memory

The system carries three memories, shown in Figure 8: a working memory holding the current dialogue, a hippocampus holding recently taught concepts, and a neocortex of consolidated facts. A sleep command, or an automatic trigger when the hippocampus fills, replays the recent traces into the neocortex and clears the fast store. A concept therefore moves from a recent, fragile state to a stable, consolidated one, and the system exposes which store answered. We probed whether the model verbalises this difference: over eight facts and three seeds, it always attributes the fact to memory rather than to pretrained knowledge (no refusals, no hallucinated source), and it tends to report a post-sleep fact as established knowledge it holds more often than a freshly taught one (71% versus 42% under a neutral register judge), but the recent versus consolidated distinction is only partial: the model often says "you taught me this earlier" in both states. To our knowledge a model reporting whether a fact is freshly taught or already consolidated has not been a focus of prior episodic-memory work for language or vision-language models, and our result suggests the internal state transition is reliable while its verbalisation is not yet sharp.

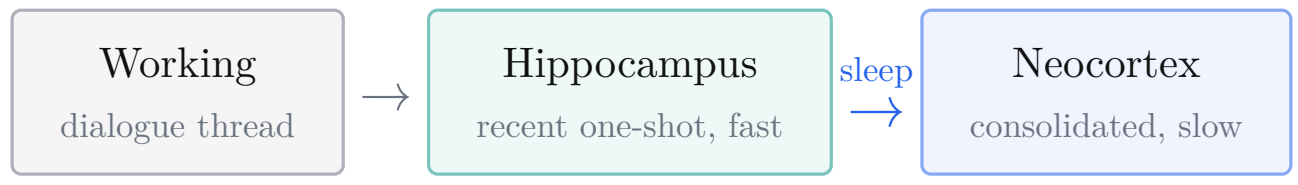

*Sleep replays the hippocampus and migrates stable memories to the neocortex (consolidation).*

Figure 8: The three-tier memory. Working memory tracks the dialogue; high-surprise inputs are written to the hippocampus; a sleep phase consolidates the hippocampus into the neocortex.

To check that consolidation actually carries facts in the slow store rather than in the dialogue context, we run a post-sleep retention test at scale. We teach the system fifty facts one-shot, trigger a sleep phase (which merges the hippocampal traces into neocortical prototypes, trains the familiarity router, and empties the hippocampus), then clear the conversation entirely and query each fact. With the hippocampus empty and no dialogue history, all fifty facts are consolidated into the neocortex and 99.2% are recalled and answered correctly (mean over five seeds, 95% CI ±0.9), while a base model stripped of in-session context recovers none (Figure 9). This isolates the slow store as the carrier of the taught knowledge: the fact survives the emptying of the fast store and the loss of conversational context. Because decoding is greedy, the seed-to-seed variance is near zero; the five seeds vary the fact sampling and order rather than the generation.

**Post-sleep recall (50 taught facts, hippocampus emptied, conversation cleared; 5 seeds)**

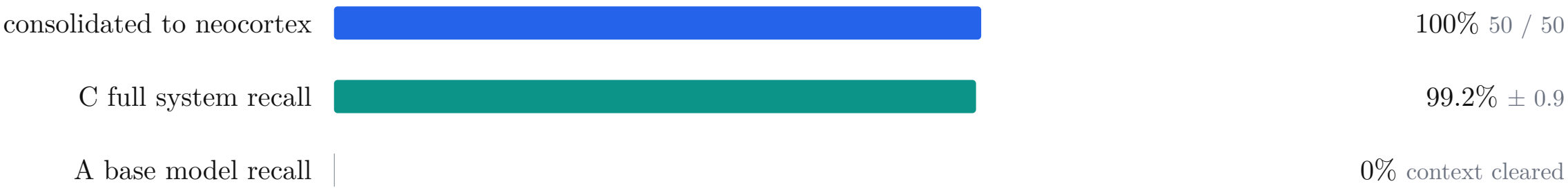

Each of fifty facts is taught once; the sleep phase consolidates it into the neocortex and trains the router, the hippocampus is emptied and the dialogue is cleared, then every fact is queried. All fifty facts survive consolidation (100%) and 99.2% are recalled (mean over five seeds, 95% CI ± 0.9). The base model, stripped of in-session context, recovers none. Greedy decoding, so the seed-to-seed variance is near zero.

Figure 9: Post-sleep retention at scale. After teaching fifty facts, sleeping, emptying the hippocampus and clearing the dialogue, all are consolidated and 99.2% are recalled from the neocortex (five seeds); the base model, without episodic memory, recovers none.

## Evaluation

We evaluate three configurations: the base vision-language model alone (A), an earlier memory design without automatic learning or text retrieval (B), and the full system (C). We collected all responses over a fixed set of cases covering fact retention, retrieval under paraphrase, respect of user corrections, and conversational memory across turns, and scored each case against a single objective pass criterion applied identically to both backbones, so the score is deterministic and does not depend on a model grading the outputs. Three of the four dimensions turned out not to separate the configurations: because the test runs within one session, the base model's native conversational context already answers the retention, paraphrased retrieval, and conversational-memory cases, and all three systems pass them. The correction dimension is the one that isolates the episodic mechanism, and Figure 10 reports it.

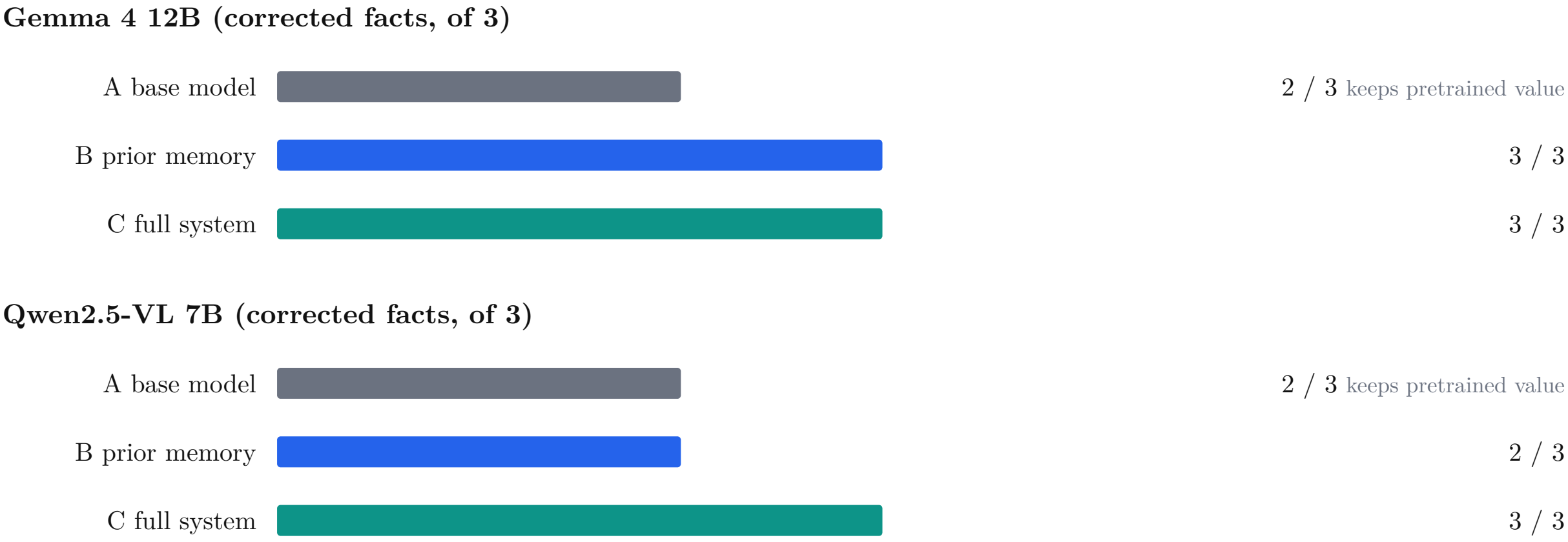

Correction task only: the one rubric dimension the base model's in-session context cannot solve, since it must override a confidently held pretrained value. Retention, paraphrased retrieval and conversational memory are satisfied by context in every configuration and do not separate them. A = base model, B = prior memory only, C = full system.

Figure 10: Correction task: number of three corrected facts answered with the corrected value rather than the model's pretrained one. Only the full system, in which the retrieved fact has authority, reaches 3/3 on both backbones.

The correction task is discriminative precisely because in-session context is not enough to pass it: the model is told a fact that contradicts a value it holds confidently from pretraining, namely that the Mont Blanc summit is 4806 m, not the 4810 m (or 4808 m in some sources) it reports by default, and must let the taught fact win. The base model reproduces its pretrained value (2 of 3 on both backbones, failing the altitude), and the earlier memory design B fixes it on Gemma but not on Qwen. Only the full system, where the retrieved fact is injected with authority over the model's prior, answers all three from the corrected fact on both backbones, attributing the answer to what it "was just taught". This is the behaviour the system is built for; the other three rubric dimensions are too easily satisfied by context to test it.

This benchmark probes the *authority* of a freshly stored fact: the cases are run in a single session without triggering a sleep cycle, so they isolate retrieval and override behaviour. The complementary weight-level consolidation, whether a taught fact survives the sleep phase and the emptying of the fast store, is measured separately by the post-sleep retention test reported above (Figure 9).

On the image side we compare the base model and the full system on one-shot recognition, discrimination, and refusal of an unknown object. The base model cannot learn a specific named object from one example and cannot tell it apart from a look-alike, so it scores 0% on both, while the full system scores 100%. On refusal of a genuinely unknown object both score 100%, since the base vision-language model is already honest when asked for a specific name it cannot know. The full pipeline runs in about 20 GB of memory with the Qwen backbone and about 30 GB with Gemma.

## Positioning

The second system sits between three families that each hold one of its pieces. The first family attaches an external signal but corrects after generation, re-ranking or filtering an answer rather than gating it beforehand. The second family abstains when uncertain, but from the model's own internal signal rather than a separate detector. The third family runs novelty or out-of-distribution detectors on a frozen vision-language encoder, but as standalone classifiers not wired into a generator. Our system uses an external frozen detector as a *pre-generation* gate, which is the intersection these families leave open.

Against the two closest personalised vision-language systems the difference is specific. MyVLM [16] does use an external recogniser, a linear classifier on frozen CLIP features, but to *activate* a known concept, not to *refuse* a novel one; it has no novelty score and no abstention, and it is few-shot rather than one-shot. Yo'LLaVA [17] places the entire decision inside the language model through learned soft tokens, with no external gate at all, and trains against hard negatives rather than refusing online. Neither verbalises a consolidation state. On the memory-consolidation side, the closest system to our first one is the sleep-consolidated memory of [20], which couples a fast key-value store with a slow semantic store and a forced or triggered offline replay, but does not clear the fast store after consolidation and does not let the model report whether a memory is recent or consolidated. The combination we report, an external surprise gate that both prevents hallucination and triggers one-shot writes, a memory that overrides the generator, and a verbalised consolidation state, does not appear as a whole in prior work, though every individual piece does.

## Limitations

Several limitations bound the claims. The System 1 results (few-shot, online continual, and the consolidation ablations) use a single seed and modest support and test sizes per class, so the reported points carry sampling noise we have not quantified with confidence intervals; the fine-grained comparisons there (for instance surprise-gated versus full replay, 84.2 versus 83.7) should be read as within single-run variation. The System 2 measurements are reported over multiple seeds with confidence intervals (eight resampled draws for the anti-hallucination AUROC, five seeds for post-sleep retention), but the benchmarks themselves are small: of the four in-session text dimensions only the correction task separates the configurations, the other three being passed by in-session context in every system, and the post-sleep retention test covers fifty facts on a single backbone, enough to establish the consolidation effect but not its scaling across backbones or fact types. The verbalisation of consolidation state is the weakest of these results: on a twenty-four-trial probe the model reliably attributes a fact to memory, but distinguishes "freshly taught" from "consolidated" only partially (71% versus 42%), and we report it as a preliminary observation rather than a measured effect. The few-shot comparison to a task-specific baseline is not like-for-like: our backbone is pretrained on large data while the baseline trains its encoder from scratch, so the comparison shows that a strong frozen backbone makes 5-way few-shot easy, not that our mechanism beats that baseline on equal footing; the 500-way regime is the more honest one. The surprise detector for the first system is a reconstruction predictor whose absolute separation is small, and the gate depends on a threshold calibrated on the base distribution; we have not tested its robustness to distribution shift in that base set. In the second system the text retrieval is deliberately conservative, so a question phrased with distant synonyms can miss a stored fact and fall back to general knowledge; this is a chosen operating point rather than a solved problem. The image side of the second system inherits the vision-language model's own perception: it grounds a concept by name through memory but does not yet enrich that memory with fine visual attributes. Our positioning against MyVLM and Yo'LLaVA is conceptual, not empirical: we compare mechanisms rather than run those systems on a shared benchmark, since our claim is about the gating behaviour they lack rather than about beating them on recognition accuracy. Finally, we keep a separate frozen SigLIP encoder in front of the speaker even when that speaker is itself a vision-language model; the detector is intentionally decoupled from the generator so that the novelty signal does not depend on the generator's own representation, which is the whole point of an external gate, but it does add a second encoder to the pipeline.

## Conclusion

We followed one signal, the prediction error of a small predictor over a frozen latent space, through two systems. As a gate it makes a non-parametric memory write only what is new, and a periodic replay phase consolidates those writes into a slow readout whose value we measured directly: on a 1000-class continual stream, consolidation recovers 17.7 points of retention for a friendly backbone and 51.3 for an unfriendly one, while a sliding-window replay is worse than none. As a modulator the same signal gives a vision-language model a usable notion of its own ignorance: it refuses to name what it has not been taught, learns a concept from one explanation, and lets that memory override its prior; it also begins to tell recent knowledge apart from consolidated knowledge, though that verbalisation is not yet reliable. The two systems share a design stance that we think is the durable point here: keep the heavy pretrained model frozen, put the novelty signal in a small separate detector, and let light, inspectable modules carry adaptation and metacognition on top. The mechanisms are simple; what proved decisive in each case was not the mechanism but the quality of the signal, the choice of metric, and the rigour of the evaluation. This is a proof of concept, not a finished system, and the limitations above mark where the next work has to go.

## Bibliography

[1] J. L. McClelland, B. L. McNaughton, and R. C. O'Reilly, “Why There Are Complementary Learning Systems in the Hippocampus and Neocortex: Insights from the Successes and Failures of Connectionist Models of Learning and Memory,” *Psychological Review*, 1995.

[2] D. Kumaran, D. Hassabis, and J. L. McClelland, “What Learning Systems do Intelligent Agents Need? Complementary Learning Systems Theory Updated,” *Trends in Cognitive Sciences*, 2016.

[3] M. Assran *et al.*, “Self-Supervised Learning from Images with a Joint-Embedding Predictive Architecture,” *CVPR*, 2023.

[4] S. Dohare, J. F. Hernandez-Garcia, Q. Lan, P. Rahman, A. R. Mahmood, and R. S. Sutton, “Loss of Plasticity in Deep Continual Learning,” *Nature*, 2024.

[5] M. Oquab *et al.*, “DINOv2: Learning Robust Visual Features without Supervision,” *TMLR*, 2024.

[6] X. Zhai, B. Mustafa, A. Kolesnikov, and L. Beyer, “Sigmoid Loss for Language Image Pre-Training,” *ICCV*, 2023.

[7] J. Chen, S. Xiao, P. Zhang, K. Luo, D. Lian, and Z. Liu, “BGE M3-Embedding: Multi-Lingual, Multi-Functionality, Multi-Granularity Text Embeddings Through Self-Knowledge Distillation,” *ACL Findings*, 2024.

[8] G. M. van de Ven, H. T. Siegelmann, and A. S. Tolias, “Brain-Inspired Replay for Continual Learning with Artificial Neural Networks,” *Nature Communications*, 2020.

[9] J. Kirkpatrick *et al.*, “Overcoming Catastrophic Forgetting in Neural Networks,” *PNAS*, 2017.

[10] A. Behrouz, P. Zhong, and V. Mirrokni, “Titans: Learning to Memorize at Test Time,” *arXiv:2501.00663*, 2025.

[11] P. Das *et al.*, “Larimar: Large Language Models with Episodic Memory Control,” *ICML*, 2024.

[12] C. Packer *et al.*, “MemGPT: Towards LLMs as Operating Systems,” *arXiv:2310.08560*, 2023.

[13] J. Snell, K. Swersky, and R. Zemel, “Prototypical Networks for Few-shot Learning,” in *NeurIPS*, 2017.

[14] O. Vinyals, C. Blundell, T. Lillicrap, D. Wierstra, and others, “Matching Networks for One Shot Learning,” in *NeurIPS*, 2016.

[15] C. Zhang, Y. Cai, G. Lin, and C. Shen, “DeepEMD: Few-Shot Image Classification with Differentiable Earth Mover's Distance and Structured Classifiers,” in *CVPR*, 2020.

[16] Y. Alaluf, E. Richardson, S. Tulyakov, K. Aberman, and D. Cohen-Or, “MyVLM: Personalizing VLMs for User-Specific Queries,” in *ECCV*, 2024.

[17] T. Nguyen, H. Liu, Y. Li, M. Cai, U. Ojha, and Y. J. Lee, “Yo'LLaVA: Your Personalized Language and Vision Assistant,” in *NeurIPS*, 2024.

[18] S. Bai, K. Chen, X. Liu, and others, “Qwen2.5-VL Technical Report,” *arXiv:2502.13923*, 2025.

[19] Gemma Team, Google DeepMind, “Gemma 4: Model Card and Technical Report.” 2026.

[20] S. S. Shinde, “SCM: Sleep-Consolidated Memory with Algorithmic Forgetting for Large Language Models,” *arXiv:2604.20943*, 2026.